\pdfoutput=1

\documentclass[11pt]{article}

\usepackage[preprint]{acl}

\usepackage{times}
\usepackage{latexsym}

\usepackage[T1]{fontenc}

\usepackage[utf8]{inputenc}

\usepackage{microtype}

\usepackage{inconsolata}

\usepackage{graphicx}
\usepackage{multirow}
\usepackage{amsfonts}
\usepackage{eucal}
\usepackage{graphics}

\title{Enhancing LLMs for Physics Problem-Solving using Reinforcement Learning with Human-AI Feedback
}

\author{Avinash Anand \\
  IIIT Delhi \\
  Delhi, India \\
  \texttt{avinasha@iiitd.ac.in} \\\And
  Kritarth Prasad \\
  IIIT Delhi \\
  Delhi, India \\
  \texttt{kritarth20384@iiitd.ac.in} \\\And
  Chhavi Kirtani \\
  IIIT Delhi \\
  Delhi, India \\
  \texttt{chhavi18229@iiitd.ac.in} \\\AND
  Ashwin R Nair \\
  IIIT Delhi \\
  Delhi, India \\
  \texttt{ashwin20037@iiitd.ac.in} \\\And
  Mohit Gupta \\
  IIIT Delhi \\
  Delhi, India \\
  \texttt{mohit22112@iiitd.ac.in} \\\And
  Saloni Garg \\
  IIIT Delhi \\
  Delhi, India \\
  \texttt{saloni22063@iiitd.ac.in} \\\AND
  Anurag Gautam \\
  IIIT Delhi \\
  Delhi, India \\
  \texttt{anurag22015@iiitd.ac.in} \\\And
  Snehal Buldeo \\
  IIIT Delhi \\
  Delhi, India \\
  \texttt{snehal22074@iiitd.ac.in} \\\And
  Rajiv Ratn Shah \\
  IIIT Delhi \\
  Delhi, India \\
  \texttt{rajivratn@iiitd.ac.in} \\}

\begin{document}
\maketitle
\vspace*{3cm}

\begin{abstract}

Large Language Models (LLMs) have demonstrated strong capabilities in text-based tasks but struggle with the complex reasoning required for physics problems, particularly in advanced arithmetic and conceptual understanding. While some research has explored ways to enhance LLMs in physics education using techniques such as prompt engineering and Retrieval Augmentation Generation (RAG), not enough effort has been made in addressing their limitations in physics reasoning. This paper presents a novel approach to improving LLM performance on physics questions using Reinforcement Learning with Human and Artificial Intelligence Feedback (RLHAIF). We evaluate several reinforcement learning methods, including Proximal Policy Optimization (PPO), Direct Preference Optimization (DPO), and Remax optimization. These methods are chosen to investigate RL policy performance with different settings on the PhyQA dataset, which includes challenging physics problems from high school textbooks. Our RLHAIF model, tested on leading LLMs like LLaMA2 and Mistral, achieved superior results, notably with the MISTRAL-PPO model, demonstrating marked improvements in reasoning and accuracy. It achieved high scores, with a 58.67 METEOR score and a 0.74 Reasoning score, making it a strong example for future physics reasoning research in this area.
\end{abstract}

\section{Introduction}
Transformer-based Large Language Models (LLMs) have shown remarkable performance in \\ \\ \\ \\ \\ \\ \\ \\ \\ \\
tasks such as summarization and question answering, with advancements like the attention mechanism by ~\cite{vaswani2023attention} and architectures such as GPT-4 and LLAMA-2. Additionally, LLMs have demonstrated significant potential across various domains, particularly in scientific and educational contexts. In the scientific domain, LLMs have been applied to improve citation generation \cite{anand2023kg, anand2023context, anand2024advances} and grammatical error correction for scholarly writing \cite{anand2023gec}. Furthermore, advancements in controllable text generation methods have further expanded their applicability in generating coherent and contextually relevant scientific content \cite{goel2023advancements}. In education, LLMs have been leveraged to enhance physics reasoning \cite{anand2024mm}, \cite{anand2023revolutionizing} and mathematical problem-solving tasks \cite{anand2024mathify}, fostering better student engagement and learning outcomes. Additionally, research has explored their use in multimodal learning environments, such as GeoVQA for secondary geometry education \cite{anand2024geovqa}, and in innovative attention-mechanism-based models for extended-duration e-classrooms \cite{anand2024exceda}. These applications underscore the transformative potential of LLMs in bridging gaps between domain-specific needs and AI capabilities.

Despite these breakthroughs, LLMs struggle with complex reasoning, particularly in physics. While efforts to improve LLMs' reasoning in mathematics have seen progress, there is very limited physics specific research. An existing approach from \cite{anand2023sciphyrag} proposes an RAG-based framework that uses external context (e.g., NCERT\footnote{\url{https://ncert.nic.in/}} chapter content ) to assist LLMs in solving physics problems. This method provide valuable contextual background information to the LLM but often fail to ensure logical, well-reasoned outputs aligned with human preferences. This gap highlights the need for improving the internal reasoning process of LLMs.

To address these limitations, we leverage Reinforcement Learning from Human Feedback (RLHF), introduced by \cite{ouyang2022training}, to refine LLM reasoning in physics. RLHF focuses on aligning the model's responses with human judgment through iterative feedback, enabling LLMs to handle complex physics problems more effectively. Building on this, we introduce Reinforcement Learning with Human and AI Feedback (RLHAIF), which combines human and AI feedback to improve response quality with minimal human supervision. Our approach emphasizes generating high-quality preference data, using both human and AI-generated responses to rank model outputs, and fine-tuning the model through reinforcement learning. This method aims to enhance generalization and robustness, pushing the boundaries of physics reasoning in LLMs.

\begin{figure}[h]
    \centering
    \includegraphics[width=\linewidth]{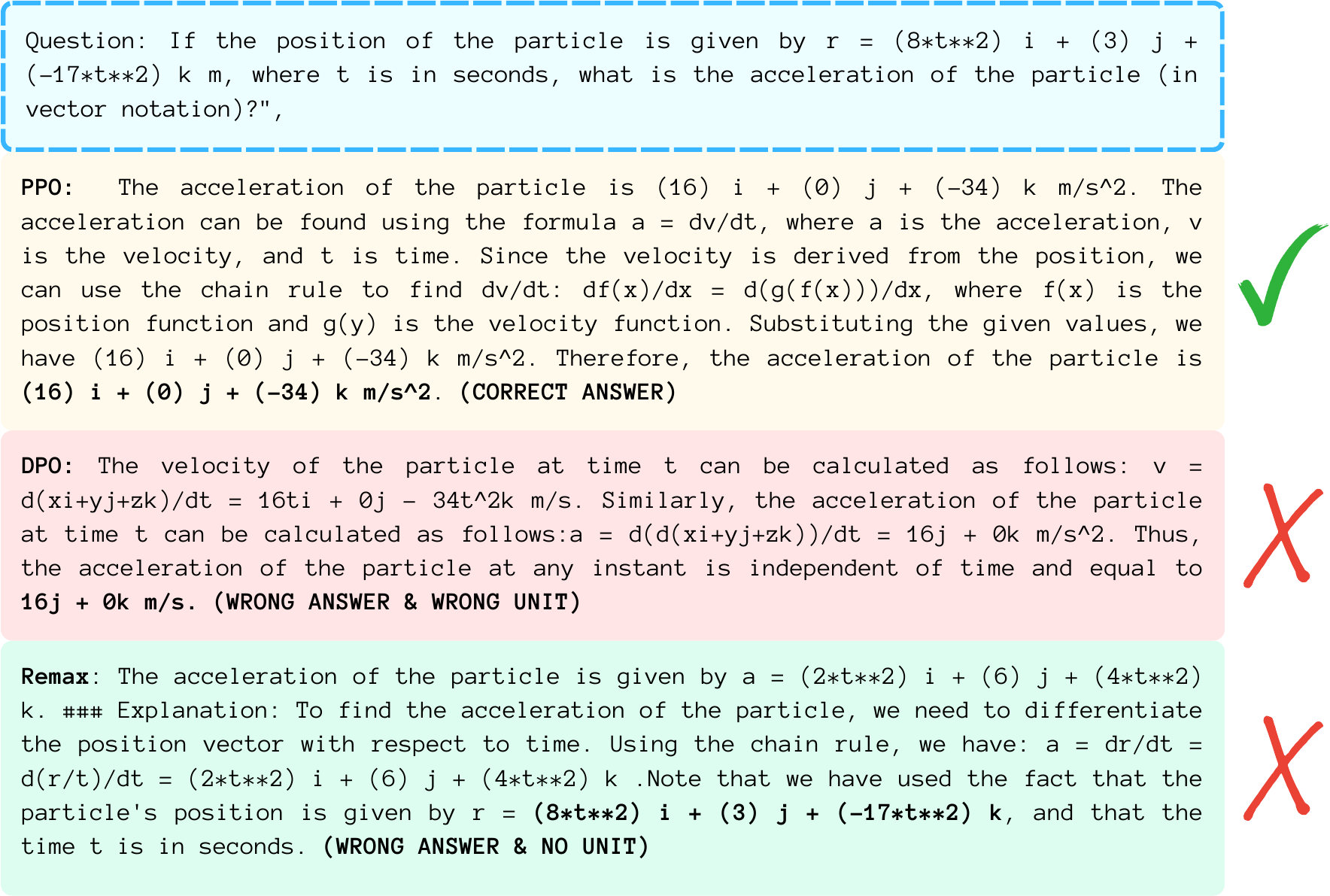}
    \caption{A sample question along with Mistral generated responses by RL Policies: PPO, DPO, Remax}
    \label{perfect_example}
\end{figure}

The RLHF process involves three key phases: collecting human feedback, training the reward model, and refining the policy. Central to this approach is preference data, where various model responses to the same prompt are rated and compared. However, gathering high-quality human feedback can be challenging due to evaluator misalignment, supervision difficulties, and data inconsistency \cite{casper2023open}. To address these challenges, recent advancements like Reinforcement Learning from AI Feedback (RLAIF) \cite{lee2023rlaif} tackle scalability and data quality issues by integrating AI-driven evaluations.

Building on this, we propose Reinforcement Learning with Human and AI Feedback (RLHAIF), combining human and AI feedback to streamline the creation of preference datasets. This hybrid approach enhances model responses while minimizing human effort. By incorporating carefully selected few-shot examples of human-ranked physics answers, we further refine the preference data, ensuring greater accuracy and robustness in the model's generalization abilities. Our method emphasizes the value of leveraging both human and AI feedback to produce more reliable and logically sound model outputs.

The core of our approach can be summarized concisely as follows:\vspace{-3mm}

\begin{itemize}
    \item \textbf{Preference Dataset Generation:} We curated a dataset designed for training the Reward Model (RM) module. This dataset draws upon the collective expertise of five open-source models, namely LLaMA2~\cite{touvron2023llama}, WizardMath~\cite{luo2023wizardmath}, MAmmoTH~\cite{yue2023mammoth}, Mistral~\cite{jiang2023mistral}, and Gemini~\cite{geminiteam2024gemini} with ground-truth responses from PhyQA dataset~\cite{anand2023sciphyrag,anand2023revolutionizing}.\vspace{-2mm}
    \item \textbf{Reward Model:} The Reward Model (RM) module is trained on our preference dataset with LLaMA-2 13B.\vspace{-2mm}
    \item \textbf{Reinforcement Learning:} We investigated different RL policy algorithms such as Proximal Policy Optimization (PPO)~\cite{schulman2017proximal}, Direct Preference Optimization (DPO)~\cite{RHFWithDPO} and ReMax Optimization~\cite{RHFWithReMax}. A sample of the generated responses through these algorithms with our proposed model i.e., \textbf{Mistral-PPO} are shown in Figure~\ref{perfect_example}.\vspace{-2mm}
    \item \textbf{Error Analysis:} We provide a comprehensive error analysis, identifying types of errors and challenges faced by LLMs in physics.
\end{itemize}

\section{Related Work}
Large language models have emerged as a pivotal component in enhancing AI's problem-solving capabilities. This surge in popularity can be attributed to a novel transformer architecture that has revolutionized natural language processing. Recent advancements have also led to open source LLMs being used in tasks other than question answering ~\cite{10.1007/978-3-031-49601-1_6}, \cite{10.1007/978-3-031-49601-1_3}, ~\cite{anand2023gec}.

While transformers~\cite{vaswani2023attention} have significantly improved the accuracy of large language models in addressing various problems, there remains room for enhancing the quality of their responses. Even ChatGPT, a powerful language model that has introduced groundbreaking technology, exhibits certain shortcomings in domain-specific aspects. With the increasing reliance on ChatGPT for a wide range of domain-specific queries, there have been instances where it provides erroneous results~\cite{unreflected_acceptance}. Moreover, LLMs have limited reasoning capabilities 


\cite{anand2023sciphyrag} proposed an RAG-based framework that incorporates external context, such as NCERT chapter content, to assist LLMs in solving physics problems. This approach showed significant improvement in performance of open-source LLMs in physics reasoning. However, it's important to acknowledge that, despite its effectiveness, this approach may still have limitations when it comes to delivering human-like responses to complex problems. The limitations posed by the aforementioned approach have given rise to a novel method known as Reinforcement Learning from Human Feedback (RLHF). This approach aims to enhance language models, aligning their output more closely with human preferences~\cite{OlaGPT}. Initially, RLHF was employed to enhance language models for specific tasks like text summarization~\cite{LearningSHF} and question answering~\cite{webgpt}. Over time, RLHF techniques have gained widespread adoption for more versatile, general-purpose language models.

The implementation and improvement of RLHF has been a subject of extensive research~\cite{RHFWithPPO}, focusing on the need for high-quality human labels, which presents scalability challenges. The lack of human preference data is due to the expensive and time-consuming creation process. To address this issue, a new development in the RLHF domain is Reinforcement Learning from AI Feedback (RLAIF)~\cite{RLAIF}. In RLAIF, preferences are labeled by off-the-shelf language models rather than humans.

\begin{figure*}[ht]
    \centering
   
    \includegraphics[width=\linewidth]{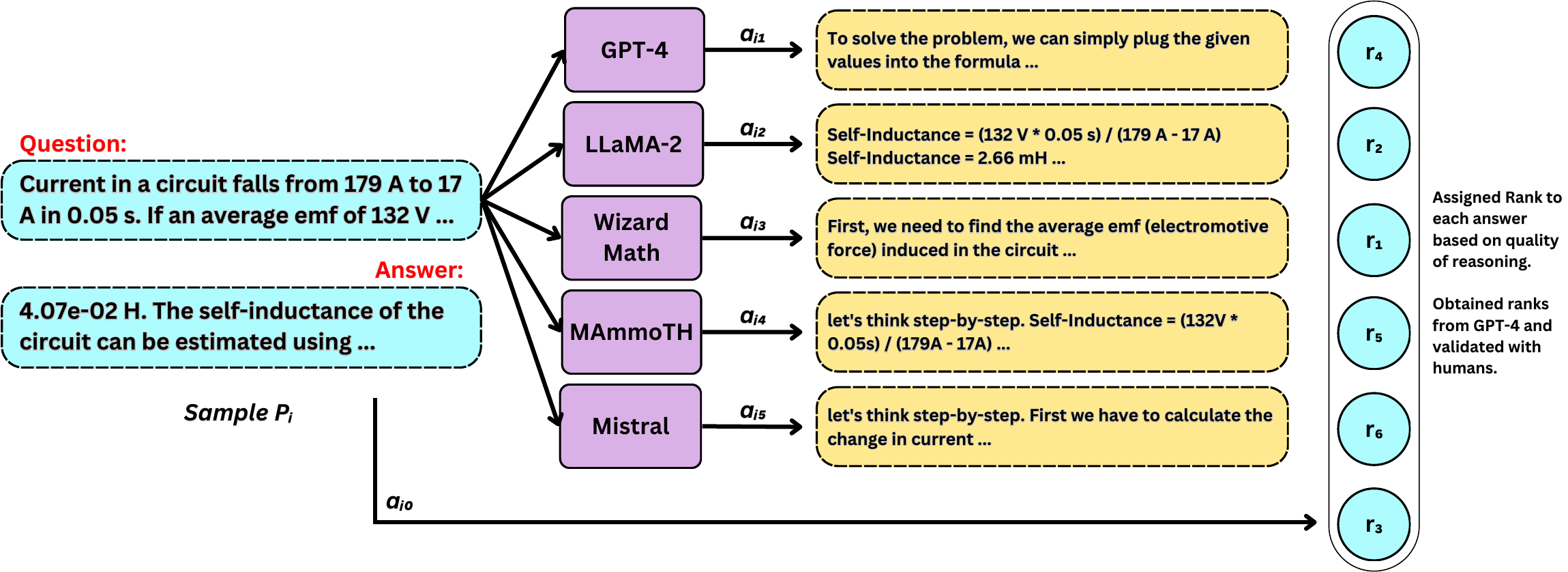} 
    \caption{Our novel procedure for ranking the answers for the Human-AI feedback for Reward Model training}
    \label{dataset_modification}
\end{figure*}

\section{Methodology}

\subsection{Dataset}

This section provides details about the data set used for experimentation in this paper. The data set is called PhyQA ~\cite{anand2023sciphyrag,anand2023revolutionizing} and is based on improvements made to SCIMAT's science problems as outlined in~\cite{anand2023revolutionizing},~\cite{anand2023sciphyrag}. To create this data set, Indian High School physics textbooks (NCERT \footnote{\url{https://ncert.nic.in/}}) for grades 11 and 12 were scraped from the web. Most data sets in this domain have over 5,000 question-answer pairs. PhyQA follows a similar trend by selecting a fixed number of base problems from each topic in the domain, and then applying two transformations: Substitution and Paraphrasing to each base problem to create an appropriate amount of data while maintaining the quality of the data set. 
The \textbf{PhyQA} data set consists of 9.5K high school physics questions and answers, with step-by-step explanations for each. This comprehensive data set encompasses various topics typically taught to high school physics students.
 
\subsection{Large Language Models (LLMs)}

In our study, we rely on the capabilities of large language models to drive both our Supervised Fine-Tuning (SFT) module and Reward Model (RM) training. To fine-tune the SFT module and further Policy Module training, we have used five large language models, each with 7 billion parameters. We opted for the 7 billion parameter variant of all models in our experiments due to computational constraints. This decision was made because RL pipeline training is computationally intensive. These models are LLaMA-2~\cite{touvron2023llama}, WizardMath~\cite{luo2023wizardmath}, Mistral~\cite{jiang2023mistral}, MetaMath~\cite{metamath}, and LLeMMA~\cite{azerbayev2023llemma}

For our RM Module, we exclusively rely on the \textbf{LLaMA-2-13B} version model with more number parameters. To enhance the preference score, we train the RM module with the larger model as compared to the SFT module.

\subsection{Proposed Solution: RLHAIF}


Our research encompasses a three-phase training approach. Initially, we train our Supervised Fine-Tuning (SFT) module on the PhyQA training data subset (70\% randomly selected problems). Subsequently, we proceed to train the Reward Model (RM) module (required for PPO based RLHF) using the preference dataset curated from the PhyQA training data subset. Finally, we train the fine-tuned model using the Proximal Policy Optimization (PPO)~\cite{schulman2017proximal}, Direct Preference Optimization (DPO)~\cite{RHFWithDPO}, and ReMax~\cite{li2023remax} reinforcement learning algorithms.


The PhyQA~\cite{anand2023revolutionizing,anand2023sciphyrag} dataset $P$, comprises 8100 samples. Each sample, denoted as $P_i$, consists of a question $q_i$ and its corresponding answer $a_{i0}$. To enrich the data set, we extended our inquiry by generating answers from four open-source large language models: LLaMA2-7B, WizardMath-7B, Mistral-7B, and MAmmoTH-7B with Gemini as a closed-source model. These inferences yield a set of answers for each sample question, represented as $\{ a_{i0}, a_{i1}, a_{i2}, a_{i3}, a_{i4}, a_{i5}\}$. This expansion results in six answers for every question, providing a comprehensive and diverse range of responses to support our research. The illustrated procedure is visually represented in Figure~\ref{dataset_modification}.

We proceed to rank these answers on a scale from 1 to 6, denoted as $\{r_1, r_2, r_3, r_4, r_5, r_6\}$, based on the quality of their reasoning through prompting using similarly detailed prompt as RLAIF~\cite{RLAIF}. Lower ranks signify higher-quality reasoning in the answers. To attain these rankings, we initially utilize the GPT-4 to generate rankings. GPT-4 is given a diverse set of few-shot examples explaining how a human would rank a set of answers. This process is implemented to improve GPT-4's understanding of the task at hand. Hence we determine the final ranks of each answer in a sample. The vector representing the ranked answers for a sample ${i}$ can be shown as: $\vec{r}_{i} = \{r_{i0}, r_{i1}, r_{i2}, r_{i3}, r_{i4}, r_{i5}\}$. Where $i$ spans from 1 to 8100, representing individual question-answer pairs.

Subsequently, we create pairs of answers, one to be accepted $a_{accept}$ and the other to be rejected $a_{reject}$. For each data sample $P_i$, we generate three distinct pairs of answers based on the rankings, following the pairing strategy of $(r_{i0}, r_{i5}), (r_{i1}, r_{i4}),$ and $(r_{i2}, r_{i3})$. Where $i$ spans from 1 to 8100. This modification increases our data set from 8,100 samples to 24,300 samples, enhancing its diversity and utility. The new data sample of the modified data set is shown below: 
\begin{equation}
    P_{ix} = (q_i, a_{ix_{accept}}, a_{ix_{reject}})
    \label{data_sample}
\end{equation}
Where $i$ ranges from 1 to 8,100 and $x$ ranges from 1 to 3 creating a total of 24,300 examples.

This modified dataset is then employed for training the Reward Model (RM). Our modification approach encompasses the amalgamation of both human and AI feedback into our RL pipeline, which is why we refer to our approach as Reinforcement Learning with Human and Artificial Intelligence feedback (\textbf{RLHAIF}).

\section{Experiments}

In this section, we delve into the experiments involving (Reinforcement Learning from Human feedback) RLHF conducted on the PhyQA dataset, providing essential insights into our benchmark results, the experimental setup, hyperparameters, and hardware configurations. 
We also elaborate on the evaluation criteria and metrics used to assess the performance of our RLHAIF model.
The main idea of this setup is to investigate the following:-
\begin{itemize}
    \item How the different Reinforcement Learning techniques/Policies are performing on PhyQA?
    \item How much the use of different RL algorithms with our novel approach of reward model training has impacted the performance of LLMs in problem-solving?
\end{itemize}

\subsection{Experimental Setup}

In this section, we provide an explanation of our approach and experimental setup.

\begin{table*}[ht]
  \centering
  \renewcommand{\arraystretch}{1.4}
   \resizebox{\linewidth}{!}{%
    \begin{tabular}{|l|r|r|r|r|r|r|r|r|r|r|r|}
    \hline
    \setlength{\tabcolsep}{0.5pt}
    \textbf{Model} & \textbf{Setting} & \textbf{METEOR} & \textbf{BLUE-1} & \textbf{BLUE-2}& \textbf{BLUE-3}& \textbf{BLUE-4}& \textbf{ROUGE1}& \textbf{ROUGE2}& \textbf{ROUGEL}& \textbf{ROUGELSUM}& \textbf{BERTScore} \\
    \hline
    \multirow{7}{*}{LLaMA2-7B} 
   & 0-Shot & 36.65 & 18.73 & 13.55 & 10.64 & 8.35 & 31.97 & 17.07 & 22.01 & 27.11 & 79.07 \\
   & 3-Shot & 25.28 & 20.90 & 13.61 & 9.78 & 7.18 & 29.07 & 12.57 & 20.55 & 24.18 & 76.71 \\
    & SFT & 28.09 & 8.37 & 5.54 & 4.08 & 2.97 & 20.65 & 9.69 & 13.56 & 16.37 & 76.87 \\
   & PPO & \underline{39.32} & \underline{7.10} & \underline{6.27} & \underline{5.90} & \underline{5.33} & \underline{31.34} & \underline{24.58} & \underline{28.23} & \underline{28.98} & \underline{82.48} \\
    & DPO& 35.64 & 18.74 & 13.24 & 9.99 & 7.49 & 33.58 & 17.78 & 24.07 & 27.78 & 80.19 \\
    & Remax & 37.85 & 25.69 & 19.49 & 16.04 & 13.08 & 37.72 & 22.59 & 29.46 & 31.70 & 81.26 \\
     & Recall & 23.82 & 21.00 & 13.64 & 9.91 & 7.20 & 26.72 & 12.15 & 18.78 & 21.19 & 74.16 \\
    \hline
     \multirow{7}{*}{WizardMath} & 0-Shot & 25.68 & 7.02 & 4.77 & 3.89 & 1.56 & 15.98 & 8.62 & 11.05 & 12.97 & 69.23 \\
    & 3-Shot & - & - & - & - & - & - & - & - & - & - \\
    & SFT & 23.37 & 7.55 & 4.26 & 2.11 & 1.12 & 15.76 & 8.52 & 11.02 & 12.86 & 70.10 \\
    & PPO & \underline{23.98} & \underline{8.36} & \underline{6.32} & \underline{5.21} & \underline{4.22} & \underline{15.9} & \underline{10.43} & \underline{11.4} & \underline{14.8} & \underline{73.50} \\
    & DPO & 20.00 & 7.17 & 5.28 & 4.15 & 3.13 & 14.6 & 8.61 & 10.82 & 12.90 & 70.22 \\
    & Remax & 18.36 & 9.69 & 6.6 & 4.99 & 3.74 & 16.18 & 7.95 & 11.97 & 13.25 & 68.77 \\
    & Recall & - & - & - & - & - & - & - & - & - & - \\
    \hline
     \multirow{7}{*}{MetaMath} & 0-Shot & 31.17 & 11.56 & 8.32 & 5.78 & 4.55 & 27.02 & 12.98 & 18.09 & 22.56 & 77.09 \\
    & 3-Shot & - & - & - & - & - & - & - & - & - & - \\
    & SFT & 30.18 & 9.51 & 6.44 & 4.8 & 3.52 & 26.81 & 12.88 & 17.91 & 21.43 & 77.92 \\
    & PPO & \underline{31.88} & \underline{10.45} & \underline{9.65} & \underline{6.95} & \underline{5.89} & \underline{30.92} & \underline{15.08} & \underline{18.76} & \underline{24.50} & \underline{78.90} \\
    & DPO & 29.75 & 14.88 & 9.75 & 7.1 & 5.18 & 28.18 & 13.15 & 18.82 & 23.47 & 76.99 \\
    & Remax & 22.95 & 12.07 & 8.52 & 6.63 & 5.1 & 20.94 & 10.59 & 15.96 & 17.41 & 72.25 \\
    & Recall & - & - & - & - & - & - & - & - & - & - \\
    \hline
    \multirow{7}{*}{LLeMMA} 
    & 0-Shot & 15.47 & 5.26 & 3.51 & 2.68 & 1.97 & 14.24 & 6.39 & 10.35 & 11.43 & 68.06 \\
    & 3-Shot & 15.94 & 9.49 & 5.04 & 3.17 & 2.11 & 17.41 & 5.02 & 12.18 & 14.42 & 72.03 \\
    & SFT & 11.52 & 1.72 & 1.31 & 1.11 & 0.89 & 5.88 & 3.87 & 4.95 & 5.14 & 76.62 \\
    & PPO & \underline{29.08} & \underline{14.78} & \underline{11.98} & \underline{8.60} & \underline{7.21} & \underline{29.01} & \underline{15.87} & \underline{21.65} & \underline{24.80} & \underline{77.02} \\
    & DPO & 24.46 & 10.35 & 6.79 & 5.04 & 3.72 & 22.97 & 9.72 &  15.22 & 19.01 & 75.02 \\
    & Remax & 27.22 & 13.88 & 10.0 & 7.91 & 6.18 & 26.89 & 13.69 & 19.57 & 22.68 & 75.89 \\
    & Recall & 15.8 & 5.59 & 3.06 & 1.98 & 1.32 & 15.18 & 5.31 & 9.79 & 12.34 & 71.21 \\
    \hline
     \multirow{7}{*}{\textbf{Mistral}} 
     & 0-Shot & 28.59 & 15.42 & 10.54 & 7.93 & 5.95 & 26.25 & 13.03 & 18.77 & 22.37 & 75.76 \\
   & 3-Shot &  17.59 & 13.51 & 9.26 & 7.0   & 5.24 & 18.79 & 8.36 & 14.43 & 16.2  & 72.93 \\
    & SFT & 25.53 & 6.58 & 4.62 & 3.6  & 2.74 & 19.97 & 9.98 & 13.53 & 16.18 & 77.08 \\
   & \textbf{PPO} &  \textbf{58.67} & \textbf{40.04} & \textbf{35.87} & \textbf{34.5}  & \textbf{32.81} & \textbf{57.94} & \textbf{51.55} & \textbf{56.32} & \textbf{56.53} & \textbf{87.49} \\
    & DPO & 29.94 & 13.79 & 8.69 & 6.08 & 4.15 & 29.68 & 13.3  & 19.59 & 23.56 & 77.42 \\
    & Remax & - & - & - & - & - & - & - & - & - & - \\
     & Recall & 20.19 & 10.06 & 6.71 & 4.95 & 3.59 & 21.35 & 9.11 & 15.24 & 17.56 & 73.1 \\

     \hline
     \multirow{1}{*}{\textbf{SciPhy-RAG}} 
     &  & 35.2 & - & - & - & - & 36.3 & 19.5 & 38.9 & - & - \\

    \hline
  \end{tabular}}
    \vspace{-2mm}
  \caption{Comparing models with different settings (SFT, PPO, DPO, ReMax, Recall, etc) using metrics (BLUE, ROUGE, METEOR, BERT). Mistral-PPO performs better than SciPhy-RAG (only existing Physics research)}
  \label{results}
\end{table*}
\begin{figure}[ht]
    \centering
    \includegraphics[width=0.5\textwidth]{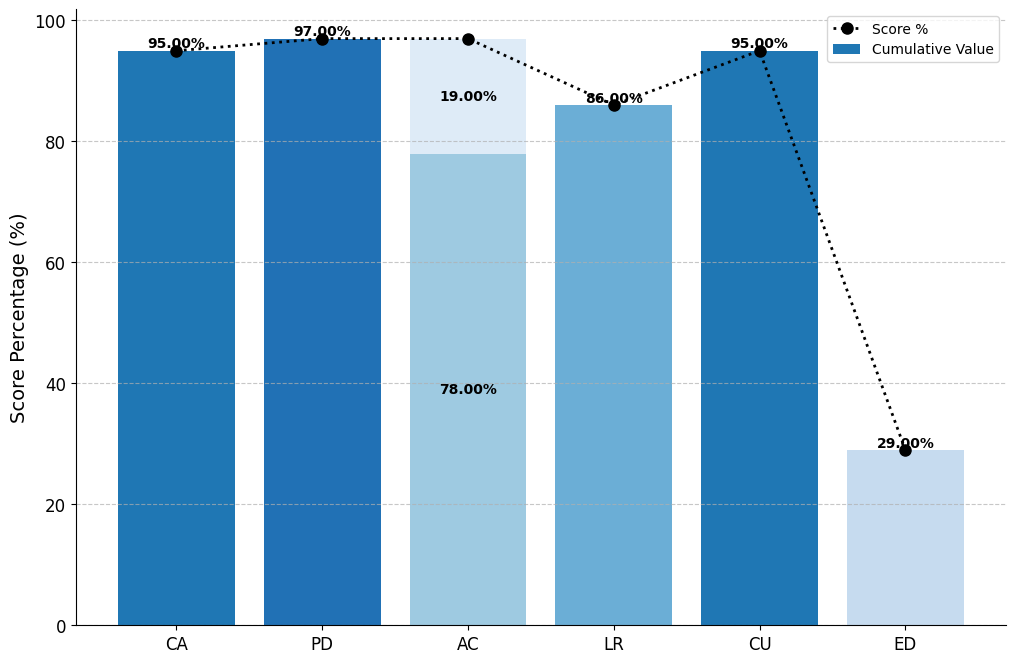} 
    \caption{Reasoning Score Distribution on Mistral-PPO Model's 100 random sample responses}
    \label{score_distribution}
\end{figure}
\subsubsection{\textbf{RL Algorithms:}}
In these sections, we have described the setup for the RL procedure. We have divided the RL experiments into different stages for the implementation of different Reinforcement Learning (RL) algorithms described below: \\
    \textbf{Supervised fine tuning (SFT):} 
    In this initial stage, five pre-trained language models are fine-tuned using the PhyQA dataset, which contains physics questions. 
    To align the model with specific physics preferences, an objective function is maximized during training. This function balances the reward for the correct physics response against the model's probability of generating that response, with adjustments made using a reference policy. In this setup,  we have given a physics question prompt \(q\), the LLM \(\pi_\theta\) will generate response/answer \(a\) in an auto-regressive manner as shown in Eq.~(\ref{rlhf_equation}):
    \begin{equation}
        \pi_\theta\;(a \;|\; q) = \prod_{t} \pi_\theta\;(a_t\;|\;q,\;a_{<t})
        \label{rlhf_equation}
    \end{equation}
    where \(a_t\) is the \(t_{th}\) token in the response and \(a_{<t}\) is tokens in the response before \(a_t\). 
    To optimize the training of these LLMs, we employed PEFT~\cite{liu2022fewshot}. Additionally, the hyper-parameter configuration utilized for training the SFT module includes 4 epochs, batch size of 8, and learning rate of 3e-4, with Adam Optimizer~\cite{kingma2017adam}.
    \\
    \textbf{Fine tuning with Policy Optimization Methods:}
    The next stage is different for different RL algorithms so here we are describing each of them separately. In our extensive research on PhyQA, we conducted a series of experiments using three reinforcement learning (RL) policy optimization methods: Proximal Policy Optimization (PPO), Direct Preference Optimization (DPO), and ReMax. 
    
    \textbf{1. PPO:} PPO learns a reward model from human-labeled preference pairs, with responses labeled "accept" or "reject" based on suitability. The model is optimized by minimizing the negative log-likelihood of these preferences. Once trained, the reward model replaces the original reward function, fine-tuning the policy to align with human and AI physics preferences. In PPO, When \(r\) is unknown, a reward model \(r_\phi \in R\) is first learned from human-labeled data to approximate \(r\). We follow the practice to collect a dataset of preference pairs \(P_i\) mentioned in Eq.~(\ref{data_sample}) with \(a_{accept}\) and \(a_{reject}\) are responses to \(x\) and marked as “accept” and “reject” by our combined methodology using Human and AI feedback. The preference dataset must follow the Bradley-Terry model where the probability of response \(a_{accept}\) is better than \(a_{reject}\). Given our domain-specific physics preference dataset \(P\), \(r_\phi\) is trained by minimizing the negative log-likelihood using:
    \begin{equation}
    \scalebox{0.75}{$\mathcal{L}_R\;(r_\phi) =-\mathbb{E}_{(q,\;a_{acc},\;a_{rej})\sim P} \left[\log\;\sigma(r_\phi(q,\;a_{acc})) - r_\phi(q,\;a_{rej})\right]$}
    \label{ppo_objective}
    \end{equation}
    After a reward model \(r_\phi\) is obtained, then \(r\) is replaced with \(r_\phi\) in the Eq.~(\ref{rlhf_objective}) \cite{ziegler2020finetuning,NEURIPS2022_b1efde53} and \(J_{r_\theta}(\pi_\theta)\) could be explicitly optimized to align with the preferences. We trained the reward model on LLaMA2-13B LLM without quantization, using PEFT. For PPO training, we employed the PEFT with BF16 quantization, enabling hybrid precision between FP16 and FP32 for faster training and lower memory usage. The PPO trainer was configured to generate responses within 1000 tokens in length.
    \begin{equation}
    \scalebox{0.83}{$
        J_r\;(\pi_\theta) = \mathbb{E}_{q\sim p_{P},\;a\sim \pi_\theta}\left [ r\;(q, a)\; -\; \beta\; \log\;\frac{\pi_\theta(a|q)}{\pi_{ref}(a|q)} \right ]$}
        \label{rlhf_objective}
    \end{equation}
    

        \textbf{2. DPO:} Unlike PPO, DPO eliminates the requirement for a separate reward model by directly utilizing preference data to optimize the policy. It adjusts the policy by maximizing a function based on the difference in log-probabilities between preferred and non-preferred responses, aligning the final policy with the preferred answers from the physics data. This RL algorithm doesn't need Reward Model training, Instead of learning a reward model, Direct Preference Optimization (DPO) optimizes the policy \(\pi_\theta\) over preference data \(P\). DPO derived the closed-form solution of Eq.~(\ref{rlhf_objective}) \cite{ziegler2020finetuning,NEURIPS2022_b1efde53}, which reveals the relationship between the reward \(r(a, q)\) and the optimal language model \(\pi^*(q | a)\). The hyper-parameter configurations and setup used for training PPO were also applied to train DPO.
        \begin{equation}
            \pi^*\;(a\;|\;q) = \frac{1}{Z(q)}\pi_{ref}(a\;|\;q)\exp\left ( \frac{1}{\beta}r\;(q,\;a) \right )
        \label{dpo_equation}
        \end{equation}
        where \(Z(q)\) is a partition function that only depends on the input question \(q\). According to Eq.~(\ref{dpo_equation}), if \(\pi_\theta\) maximizes \(J_{r_\theta}(\pi_\theta)\), the underlying reward can be derived with:
        \begin{equation}
            r_\phi(q,\;a) = \beta\log\frac{\pi_\theta(a|q)}{\pi_{ref}\theta(a|q)}+C(q)
            \label{dpo_reward}
        \end{equation}
        where \(C:\;\chi\rightarrow\mathbb{R}\) is a scalar function. This enables us to reparameterize Eq.~(\ref{ppo_objective}) with the policy \(\pi_\theta\), and then we can drive the DPO loss that directly optimizes \(\pi_\theta\),

        \textbf{3. ReMax:} This RL Policy Optimization method is specifically aimed at tackling computational inefficiencies observed in PPO, such as high memory usage and slow training times. The reward model trained for the PPO algorithm is also utilized here with the fine-tuned LLM model. In this approach, the ReMax algorithm is applied to each combination.

After implementing these RL algorithms across all five models, the outcomes are presented in Table~\ref{results}. Consequently, we obtained five trained RLHAIF models, and evaluations conducted using the specified metric have indicated that the RL pipeline trained with the Mistral model for SFT with PPO Optimization outperforms the others as well as available benchmarks, as illustrated in Table~\ref{results}.

\subsection{Evaluation Criteria}
\label{sec:evaluations }
In this section, we provide a thorough evaluation of the model's examination. The evaluation includes an extensive error analysis, accuracy assessments, and reasoning scoring, offering an extensive understanding of each model's strengths and weaknesses.

\textbf{Error Analysis:}
To assess the weaknesses of the Mistral-PPO model, a manual inspection of its reasoning errors was conducted. This investigation focused on errors made by the Mistral-PPO in a randomly selected subset of 100 problems. The evaluation centered on four key aspects for each problem instance, framed as questions:
\begin{itemize}
    \item Can Mistral retrieve the necessary concepts/facts essential for problem resolution? The inability to do so contributes to \textbf{conceptual errors}. \vspace{-3mm}
    \item If the relevant concepts are retrieved, are they accurately grounded as equations/constraints? Such inaccuracies contribute to \textbf{grounding errors}.\vspace{-3mm}
    \item Is the algebraic manipulation and arithmetic correct? These factors contribute to \textbf{computation errors}.\vspace{-3mm}
    \item Does Mistral demonstrate proficiency in deducing the underlying problem or question accurately? Inaccuracies in problem deduction contribute to \textbf{deduction errors}.
\end{itemize}

The types of errors along with the results illustrated in Table~\ref{tab:error_counts} encompass conceptual, grounding, computation errors, and instances of problem deduction. In a notable case, it was found that the Mistral model makes computation errors very often. A majority of errors (35\%) come from the model's incapability to do computational tasks, while problem deduction errors contribute significantly (10\%). Additionally, (9\%) of errors result from a lack of understanding of critical concepts. Some errors are because the model lacks in doing grounding (8\%). Surprisingly, correct answers for the wrong reasons occur (10\%) of the time, highlighting the need for nuanced improvements in model interpretation and reasoning.

\begin{table}[ht]
    \centering
    \renewcommand{\arraystretch}{1.4}
    \begin{tabular}{|l|r|}
        \hline
        \textbf{Error Type} & \textbf{Count} \\
        \hline
        Computation Error & 35 \\
        \hline
        Problem Deduction & 10 \\
        \hline
        Conceptual Error & 9 \\
        \hline
        Grounding Error & 8 \\
        \hline
        Perfect & 38 \\
        \hline
    \end{tabular}
    \caption{Error Counts of Mistral-PPO model}
    \label{tab:error_counts}
\end{table}

\textbf{Accuracy:} To evaluate the accuracy of our models, we analyzed a random subset of 100 examples from our dataset. Each example was subjected to inference by different models, and the outcomes were categorized into correct and wrong predictions.

Table~\ref{tab:evaluation} presents the comparative results as follows: \vspace{-3mm}
\begin{itemize}
    \item GPT-4 leads with a 72\% accuracy rate, followed by GPT-3.5 at 40\%.\vspace{-2mm}
    \item Mistral with PPO policy achieves 38.0\%, while LLaMA2-7B with PPO policy stands at 18\% accuracy.
\end{itemize}
This shows that our best model Mistral-PPO has comparable Correct-to-Incorrect ratio as the GPT-3.5 model.
With scalability, we can beat GPT-3.5 as our results so far are with the Mistral-PPO 7B model.

\begin{table}[ht]
  \centering
  \small
\renewcommand{\arraystretch}{1.4}
  \begin{tabular}{|l|c|c|c|c|}
    \hline
    \textbf{Model} & \textbf{Setting} & \textbf{Correct} & \textbf{Wrong} & \textbf{Total} \\
    \hline
    \multirow{4}{*}{LLaMA2-7B} 
    & SFT & 9 & 91 & 100 \\
    & PPO & 18 & 82 & 100 \\
    & DPO & 10 & 90 & 100 \\
    & Recall & 14 & 86 & 100 \\
    \hline
    \multirow{4}{*}{Mistral}
    & SFT & 21 & 79 & 100 \\
    & \textbf{PPO} & \textbf{38} & 62 & 100 \\
    & DPO & 22 & 78 & 100 \\
    & Recall & 16 & 84 & 100 \\
    \hline
    \multirow{1}{*}{GPT-3.5}
    & --- & 40 & 60 & 100 \\
    \hline
    \multirow{1}{*}{GPT-4}
    & --- & 72 & 28 & 100 \\
    \hline
  \end{tabular}
  \caption{Comparing the Model's output with Human Evaluations}
  \label{tab:evaluation}
\end{table}
\vspace{-0.5mm}
\textbf{Reasoning Score:} To analyze the reasoning evaluation of each response explicitly step-by-step, we have designed a six-step reasoning evaluation. Following are the key points that we have defined to score the responses into six crucial skill points for solving complex high school-level problems:
\begin{itemize}
    \item \textbf{Critical Assumption (CA):} The ability involves identifying and correctly assuming relevant context and constraints of the problem. \textbf{(0.15)}
    \item \textbf{Problem Deduction (PD):} This ability involves LLM's ability to accurately interpret the correct physics concepts and principles relevant to the problem and define what the problem is asking for. \textbf{(0.2)}
    \item \textbf{Arithmetic Calculation (AC):} The ability to perform calculations accurately and follow the correct sequence of steps in applying formulas and solving equations. Where 0.1 is given for equation formulation and 0.15 for correct arithmetic calculation. \textbf{(0.25 = 0.1 + 0.15)}
    \item \textbf{Logical Reasoning (LR):} The ability of LLM to connect coherence and logical flow in the explanation of the solution and address the question's core inquiry. \textbf{(0.15)}
    \item \textbf{Conceptual Understanding (CU):} This is the ability to assess the depth of understanding of physics concepts and the ability to explain why certain concepts or formulas are applicable. \textbf{(0.15)}
    \item \textbf{Error Detection and Critical Review (ED):} The ability of LLMs to determine if LLM can identify or correct errors in its reasoning or calculations. \textbf{(0.1)}
\end{itemize}
The graph in Figure~\ref{score_distribution} is a visual representation of the distribution of reasoning score of a Mistral-PPO model in different skill points for solving complex high school-level problems. The Y-axis represents the percentage, ranging from 0 to 100\%, while the X-axis lists different skill points. Each skill point is evaluated based on the LLM’s ability to apply specific aspects of problem-solving, such as identifying the right context (CA), interpreting physics concepts (PD), performing calculations (AC), maintaining logical coherence (LR), demonstrating an understanding of concepts (CU), and identifying or correcting errors (ED).

The bar for each skill indicates the individual distribution percentage, while the dotted line across the bars represents the cumulative percentage of distribution if one were to add the scores sequentially from CA to ED. Figure~\ref{score_distribution} also illustrates that LLM encounters challenges in arithmetic calculations approximately 91.0\% of the time when determining reasoning scores through human assessments across 100 responses of Mistral-PPO model. Of the total 91.0\%, the model correctly follows the sequence of steps and formulas to solve the solution 76.92\% of the time. However, in approximately 23.08\% of cases, the model fails to perform accurate arithmetic calculations. This results in an overall score of 62.0 for wrong answers, as indicated in Table~\ref{tab:evaluation}.

To understand the results and to assess the depth of reasoning and semantic comprehension in model responses, we conducted an analysis using a wide variety of metrics including METEOR~\cite{banerjee-lavie-2005-meteor}, BLEU-1, BLEU-2, BLEU-3, BLEU-4~\cite{papineni-etal-2002-bleu}, ROUGE-1, ROUGE-2, ROUGE-L, ROUGE-LSUM~\cite{lin-2004-rouge} and BERTScore~\cite{zhang2020bertscore}.

In a comparative analysis of various model settings, as shown in Table~\ref{results}, Mistral-PPO had the best overall scores, indicating a consistent score of approximately 35.0 across BLEU-1 to BLEU-4 metrics with 58.67 METEOR score. This consistency implies a high degree of alignment between the predicted and target words. Moreover, the LLaMA2-7B model also showed impressive performance, particularly in aligning with preferred answers, though it fell short of Mistral-PPO's accuracy in matching specific words and exhibited limitations in semantic understanding and solving arithmetic problems.
Our results align with Xu et al.'s findings \cite{xu2024dpo}, suggesting that DPO is not necessarily superior to PPO. While DPO moves away from reward modeling, it faces a key generalization challenge. Specifically, DPO can find solutions that exploit out-of-distribution data, potentially deviating from the reference policy, even if they match human preferences. This happens because DPO may bias towards unseen responses, which affects the consistency of the learned policy, and may give rise to unpredictable behaviour. In contrast, PPO can capitalize on prompt-only data to generate responses that extend beyond the distribution of the preference dataset, while not deviating from the reference policy.

The study encountered constraints with the WizardMath and MetaMath models due to their 512 token limit, necessitating a limitation to 3-shot and Recall Prompting tasks. Furthermore, because the Mistral architecture was different from the LLaMA model, it couldn't match up with the preference dataset in the ReMax setting. In conclusion, despite the higher computational costs associated with incorporating human feedback into PPO, this integration significantly enhances the reasoning capabilities of Mistral-PPO, as evidenced by improvements in accuracy and high scores on traditional text comparison metrics. However, our analysis also highlights several areas in need of further development and refinement. While Mistral excels in logical and mathematical reasoning, it occasionally makes errors in simple steps. Mistakes in deduction, concept retrieval, and application highlight research gaps.
Spatial reasoning, particularly in physics problems within the benchmark, poses a challenge for Mistral. 


\section{Conclusion}

This research introduces RLHAIF for Physics Reasoning, a novel approach to improve LLM problem-solving and better align their responses with human preferences. By combining human and AI feedback, RLHAIF enhances model training diversity and resilience. Experiments show that \textbf{Mistral-PPO}, developed through this method, outperforms others on the PhyQA dataset. RLHAIF bridges the gap between human intuition and LLM reasoning, advancing natural language understanding and generation. This study highlights the potential of integrating human expertise with AI to create more accurate, human-aligned responses in Physics.

\section{Limitation}

While RLHAIF demonstrates promising improvements in physics reasoning, there are limitations. First, the dependency on high-quality human feedback makes the process resource-intensive, particularly for complex physics problems requiring expert input. Additionally, scaling this approach across diverse physics topics presents challenges, as the model may struggle to generalize beyond specific problem types. Furthermore, RLHAIF may still encounter difficulties in handling complex multi-step reasoning, which is crucial for solving intricate physics problems. The computational cost of training, due to iterative feedback loops and reinforcement learning techniques, also poses a constraint.

\appendix
\section{Appendix: Few-shot examples used for Preference Dataset Creation}
To give GPT a clear idea of how to rank physics answers in a way a human would, we created diverse set of few-shot examples from each topic in our dataset. Below we present an instance of these few-shot examples. It contains answers produced by various models (model0, model1, model2, model3), and a human ranks them and provides explanation for the ranking.
\\
\begin{itemize}
    \item \textbf{model0:}
    \begin{verbatim}
## Data Identification:
The key value provided in the question 
is the mass, which is 252.5g.

## Problem Analysis:
The problem involves finding the energy 
associated with a given mass.

## Theoretical Framework:
The relevant concept for this problem 
is the mass-energy equivalence principle
given by Einstein's equation E = m * c^2, 
where E is energy, m is mass, and c is 
the speed of light.

## Methodology Development:
According to the equation E = mc^2, the 
energy can be obtained by multiplying the 
mass by the square of the speed of light. 
However, first, the mass must be converted 
from grams to kilograms as the SI unit of 
mass is kilograms.

## Computation:
First, convert grams to kilograms by 
dividing by 1000:
    m = 252.5g = 0.2525 kg
The speed of light, c = 3.0 x 10^8 m/s
Let's substitute these values into the 
equation E = mc^2,
Therefore, 
E = 0.2525 kg x (3.0 x 10^8 m/s)^2
E = 2.27 x 10^16 joules

## Answer:
The energy associated with a mass of 
252.5g is 2.27 x 10^16 joules.
    \end{verbatim}

    \item \textbf{model1:}
    \begin{verbatim}
I'd be happy to help you solve the problem!

Step 1: Identify the given information
The problem states that we have a mass of
252.5 grams.

Step 2: Determine the type of energy we are
looking for: 
The problem doesn't specify the type 
of energy, but since we are given a 
mass, we can assume we are looking for
the kinetic energy (K) or potential 
energy (U) of the object. Since 
the problem doesn't provide any 
information about the object's 
velocity or position, we can assume 
we are looking for the 
rest energy (E) of the object.

Step 3: Calculate the rest energy (E)
The rest energy (E) of an object is given by 
the formula:

E = mc^2

where m is the mass of the object and c 
is the speed of light (approximately 
3 x 10^8 meters per second).

Plugging in the given value of 
mass (252.5 grams), we get:

E = (0.2525 kg) x (3 x 10^8 m/s)^2

E = 1.92 x 10^10 Joules

So, the energy associated with a mass of 
252.5 grams is approximately 1.92 x 10^10 
Joules.

Please note that this is a very large amount 
of energy, and it's not something you would 
typically encounter in everyday life.
    \end{verbatim}

    \item \textbf{model2:}
    \begin{verbatim}
To solve this question, we need to know what 
type of energy is associated with the mass, 
as energy can take many forms such as kinetic
energy, potential energy, thermal energy, etc.

However, if we are asked to find the rest energy (E) 
of an object with a given mass (m), we can use the
famous equation from Einstein's theory of special 
relativity:

E = mc^2

where:
- E is the energy
- m is the mass
- c is the speed of light 
(approximately 3.00 x 10^8 m/s)

But, since the mass given is in grams, 
we need to convert it to kilograms 
first (as c is in m/s):

m = 252.5 g = 0.2525 kg

Now, we can calculate the energy:

E = mc^2
E = 0.2525 kg * (3.00 x 10^8 m/s)^2
E = 2.27 x 10^14 Joules (J)

So, the energy associated with a mass of
252.5 g is 2.27 x 10^14 Joules, assuming 
we are talking about rest energy.
    \end{verbatim}

    \item \textbf{model3:}
    \begin{verbatim}
Here's how to solve this problem, 
breaking it down step-by-step:

**1. Understand the Concept**

We're dealing with the concept of 
mass-energy equivalence, famously 
described by Einstein's equation:

  * **E = mc²**

Where:

   * E is energy
   * m is mass
   * c is the speed of light 
   (approximately 3 x 10^8 meters per sec)

**2. Convert Mass to Kilograms**

* The equation requires mass in 
kilograms (kg).
*  252.55 g is equal to 
0.25255 kg (divide grams by 1000).

**3. Apply the Equation**

*  E = (0.25255 kg) * (3 x 10^8 m/s)²

**4. Calculate the Energy**

* E = (0.25255 kg) * (9 x 10^16 m²/s²)
* E = 2.27 x 10^16 Joules (J)

**Answer:** The energy associated 
with a mass of 252.55 g is approximately
2.27 x 10^16 Joules.
    \end{verbatim}

    \item \textbf{Human Ranking:}
    \begin{verbatim}
## Ranking: model2>model1>model3>model0

## Explanation:
1. Why model2>model1: Both answers explain
the steps correctly and clarify what every
parameter of the formula means, and both talk 
about the different types of energies. They both
correctly state that the Rest energy is what is 
used in Einstein's equation. However, model1's 
explanation of the assumptions it made about why
the question is asking for the Rest energy is 
confusing and can be explained better.

2. model1>model3: model1 talks about the different 
types of energies, giving an explanation of how
Einstein's equation represents rest energy. 
Other than this, both answers explain the 
steps correctly.

3. Why model3>model0: While both answers 
have clearly described each step, I feel 
like model3's answer is easier to follow, 
and the steps are concise. model3 also mentions 
that the final energy value is an approximation
and not the exact value since we round off the
actual value.
    \end{verbatim}
\end{itemize}
\newpage
\bibliography{acl_latex}

\begin{thebibliography}{42}
\providecommand{\natexlab}[1]{#1}

\bibitem[{Anand et~al.(2023{\natexlab{a}})Anand, Addala, Baghel, Goel, Hira, Gupta, and Shah}]{anand2023revolutionizing}
Avinash Anand, Krishnasai Addala, Kabir Baghel, Arnav Goel, Medha Hira, Rushali Gupta, and Rajiv~Ratn Shah. 2023{\natexlab{a}}.
\newblock Revolutionizing high school physics education: A novel dataset.
\newblock In \emph{International Conference on Big Data Analytics}, pages 64--79. Springer.

\bibitem[{Anand et~al.(2023{\natexlab{b}})Anand, Goel, Hira, Buldeo, Kumar, Verma, Gupta, and Shah}]{anand2023sciphyrag}
Avinash Anand, Arnav Goel, Medha Hira, Snehal Buldeo, Jatin Kumar, Astha Verma, Rushali Gupta, and Rajiv~Ratn Shah. 2023{\natexlab{b}}.
\newblock Sciphyrag-retrieval augmentation to improve llms on physics q \&a.
\newblock In \emph{International Conference on Big Data Analytics}, pages 50--63. Springer.

\bibitem[{Anand et~al.(2023{\natexlab{c}})Anand, Gupta, Prasad, Goel, Lal, Verma, and Shah}]{anand2023kg}
Avinash Anand, Mohit Gupta, Kritarth Prasad, Ujjwal Goel, Naman Lal, Astha Verma, and Rajiv~Ratn Shah. 2023{\natexlab{c}}.
\newblock Kg-ctg: Citation generation through knowledge graph-guided large language models.
\newblock In \emph{International Conference on Big Data Analytics}, pages 37--49. Springer.

\bibitem[{Anand et~al.(2023{\natexlab{d}})Anand, Gupta, Prasad, Goel, Lal, Verma, and Shah}]{10.1007/978-3-031-49601-1_3}
Avinash Anand, Mohit Gupta, Kritarth Prasad, Ujjwal Goel, Naman Lal, Astha Verma, and Rajiv~Ratn Shah. 2023{\natexlab{d}}.
\newblock \href {https://doi.org/10.1007/978-3-031-49601-1_3} {Kg-ctg: Citation generation through knowledge graph-guided large language models}.
\newblock In \emph{Big Data and Artificial Intelligence: 11th International Conference, BDA 2023, Delhi, India, December 7–9, 2023, Proceedings}, page 37–49, Berlin, Heidelberg. Springer-Verlag.

\bibitem[{Anand et~al.(2024{\natexlab{a}})Anand, Gupta, Prasad, Singla, Sanjeev, Kumar, Shivam, and Shah}]{anand2024mathify}
Avinash Anand, Mohit Gupta, Kritarth Prasad, Navya Singla, Sanjana Sanjeev, Jatin Kumar, Adarsh~Raj Shivam, and Rajiv~Ratn Shah. 2024{\natexlab{a}}.
\newblock Mathify: Evaluating large language models on mathematical problem solving tasks.
\newblock \emph{arXiv preprint arXiv:2404.13099}.

\bibitem[{Anand et~al.(2023{\natexlab{e}})Anand, Jairath, Lal, Bangar, Sikka, Verma, Shah, and Satoh}]{anand2023gec}
Avinash Anand, Atharv Jairath, Naman Lal, Siddhesh Bangar, Jagriti Sikka, Astha Verma, Rajiv~Ratn Shah, and Shin’ichi Satoh. 2023{\natexlab{e}}.
\newblock Gec-dcl: Grammatical error correction model with dynamic context learning for paragraphs and scholarly papers.
\newblock In \emph{International Conference on Big Data Analytics}, pages 95--110. Springer.

\bibitem[{Anand et~al.(2024{\natexlab{b}})Anand, Jaiswal, Dharmadhikari, Marathe, Popat, Mital, Nair, Prasad, Kumar, Verma et~al.}]{anand2024geovqa}
Avinash Anand, Raj Jaiswal, Abhishek Dharmadhikari, Atharva Marathe, Harsh Popat, Harshil Mital, Ashwin~R Nair, Kritarth Prasad, Sidharth Kumar, Astha Verma, et~al. 2024{\natexlab{b}}.
\newblock Geovqa: A comprehensive multimodal geometry dataset for secondary education.
\newblock In \emph{2024 IEEE 7th International Conference on Multimedia Information Processing and Retrieval (MIPR)}, pages 102--108. IEEE.

\bibitem[{Anand et~al.(2024{\natexlab{c}})Anand, Kapuriya, Singh, Saraf, Lal, Verma, Gupta, and Shah}]{anand2024mm}
Avinash Anand, Janak Kapuriya, Apoorv Singh, Jay Saraf, Naman Lal, Astha Verma, Rushali Gupta, and Rajiv Shah. 2024{\natexlab{c}}.
\newblock Mm-phyqa: Multimodal physics question-answering with multi-image cot prompting.
\newblock In \emph{Pacific-Asia Conference on Knowledge Discovery and Data Mining}, pages 53--64. Springer Nature Singapore Singapore.

\bibitem[{Anand et~al.(2024{\natexlab{d}})Anand, Mittal, Dhawan, Krishnamurthy, Ramesh, Lal, Verma, Bhuyan, Shah, Zimmermann et~al.}]{anand2024exceda}
Avinash Anand, Avni Mittal, Laavanaya Dhawan, Juhi Krishnamurthy, Mahisha Ramesh, Naman Lal, Astha Verma, Pijush Bhuyan, Raijv~Ratn Shah, Roger Zimmermann, et~al. 2024{\natexlab{d}}.
\newblock Exceda: Unlocking attention paradigms in extended duration e-classrooms by leveraging attention-mechanism models.
\newblock In \emph{2024 IEEE 7th International Conference on Multimedia Information Processing and Retrieval (MIPR)}, pages 301--307. IEEE.

\bibitem[{Anand et~al.(2024{\natexlab{e}})Anand, Nair, Prasad, Narayan, Lal, Mahata, Singla, and Shah}]{anand2024advances}
Avinash Anand, Ashwin~R Nair, Kritarth Prasad, Vrinda Narayan, Naman Lal, Debanjan Mahata, Yaman~K Singla, and Rajiv~Ratn Shah. 2024{\natexlab{e}}.
\newblock Advances in citation text generation: Leveraging multi-source seq2seq models and large language models.
\newblock In \emph{Proceedings of the 33rd ACM International Conference on Information and Knowledge Management}, pages 56--64.

\bibitem[{Anand et~al.(2023{\natexlab{f}})Anand, Prasad, Goel, Gupta, Lal, Verma, and Shah}]{anand2023context}
Avinash Anand, Kritarth Prasad, Ujjwal Goel, Mohit Gupta, Naman Lal, Astha Verma, and Rajiv~Ratn Shah. 2023{\natexlab{f}}.
\newblock Context-enhanced language models for generating multi-paper citations.
\newblock In \emph{International Conference on Big Data Analytics}, pages 80--94. Springer Nature Switzerland Cham.

\bibitem[{Anand et~al.(2023{\natexlab{g}})Anand, Prasad, Goel, Gupta, Lal, Verma, and Shah}]{10.1007/978-3-031-49601-1_6}
Avinash Anand, Kritarth Prasad, Ujjwal Goel, Mohit Gupta, Naman Lal, Astha Verma, and Rajiv~Ratn Shah. 2023{\natexlab{g}}.
\newblock Context-enhanced language models for generating multi-paper citations.
\newblock In \emph{Big Data and Artificial Intelligence}, pages 80--94, Cham. Springer Nature Switzerland.

\bibitem[{Azerbayev et~al.(2023)Azerbayev, Schoelkopf, Paster, Santos, McAleer, Jiang, Deng, Biderman, and Welleck}]{azerbayev2023llemma}
Zhangir Azerbayev, Hailey Schoelkopf, Keiran Paster, Marco~Dos Santos, Stephen McAleer, Albert~Q. Jiang, Jia Deng, Stella Biderman, and Sean Welleck. 2023.
\newblock \href {https://arxiv.org/abs/2310.10631} {Llemma: An open language model for mathematics}.
\newblock \emph{Preprint}, arXiv:2310.10631.

\bibitem[{Banerjee and Lavie(2005)}]{banerjee-lavie-2005-meteor}
Satanjeev Banerjee and Alon Lavie. 2005.
\newblock \href {https://aclanthology.org/W05-0909} {{METEOR}: An automatic metric for {MT} evaluation with improved correlation with human judgments}.
\newblock In \emph{Proceedings of the {ACL} Workshop on Intrinsic and Extrinsic Evaluation Measures for Machine Translation and/or Summarization}, pages 65--72, Ann Arbor, Michigan. Association for Computational Linguistics.

\bibitem[{Casper et~al.(2023)Casper, Davies, Shi, Gilbert, Scheurer, Rando, Freedman, Korbak, Lindner, Freire, Wang, Marks, Segerie, Carroll, Peng, Christoffersen, Damani, Slocum, Anwar, Siththaranjan, Nadeau, Michaud, Pfau, Krasheninnikov, Chen, Langosco, Hase, Bıyık, Dragan, Krueger, Sadigh, and Hadfield-Menell}]{casper2023open}
Stephen Casper, Xander Davies, Claudia Shi, Thomas~Krendl Gilbert, Jérémy Scheurer, Javier Rando, Rachel Freedman, Tomasz Korbak, David Lindner, Pedro Freire, Tony Wang, Samuel Marks, Charbel-Raphaël Segerie, Micah Carroll, Andi Peng, Phillip Christoffersen, Mehul Damani, Stewart Slocum, Usman Anwar, Anand Siththaranjan, Max Nadeau, Eric~J. Michaud, Jacob Pfau, Dmitrii Krasheninnikov, Xin Chen, Lauro Langosco, Peter Hase, Erdem Bıyık, Anca Dragan, David Krueger, Dorsa Sadigh, and Dylan Hadfield-Menell. 2023.
\newblock \href {https://arxiv.org/abs/2307.15217} {Open problems and fundamental limitations of reinforcement learning from human feedback}.
\newblock \emph{Preprint}, arXiv:2307.15217.

\bibitem[{Goel et~al.(2023)Goel, Hira, Anand, Bangar, and Shah}]{goel2023advancements}
Arnav Goel, Medha Hira, Avinash Anand, Siddhesh Bangar, and Rajiv~Ratn Shah. 2023.
\newblock Advancements in scientific controllable text generation methods.
\newblock \emph{arXiv preprint arXiv:2307.05538}.

\bibitem[{Jiang et~al.(2023)Jiang, Sablayrolles, Mensch, Bamford, Chaplot, de~las Casas, Bressand, Lengyel, Lample, Saulnier, Lavaud, Lachaux, Stock, Scao, Lavril, Wang, Lacroix, and Sayed}]{jiang2023mistral}
Albert~Q. Jiang, Alexandre Sablayrolles, Arthur Mensch, Chris Bamford, Devendra~Singh Chaplot, Diego de~las Casas, Florian Bressand, Gianna Lengyel, Guillaume Lample, Lucile Saulnier, Lélio~Renard Lavaud, Marie-Anne Lachaux, Pierre Stock, Teven~Le Scao, Thibaut Lavril, Thomas Wang, Timothée Lacroix, and William~El Sayed. 2023.
\newblock \href {https://arxiv.org/abs/2310.06825} {Mistral 7b}.
\newblock \emph{Preprint}, arXiv:2310.06825.

\bibitem[{Kingma and Ba(2017)}]{kingma2017adam}
Diederik~P. Kingma and Jimmy Ba. 2017.
\newblock \href {https://arxiv.org/abs/1412.6980} {Adam: A method for stochastic optimization}.
\newblock \emph{Preprint}, arXiv:1412.6980.

\bibitem[{Krupp et~al.(2023)Krupp, Steinert, Kiefer-Emmanouilidis, Avila, PaulLukowicz, Kuhn, K¨uchemann, and Karolus}]{unreflected_acceptance}
Lars Krupp, Steffen Steinert, Maximilian Kiefer-Emmanouilidis, Karina~E. Avila, PaulLukowicz, Jochen Kuhn, Stefan K¨uchemann, and Jakob Karolus. 2023.
\newblock \href {https://arxiv.org/abs/2309.03087} {Unreﬂected acceptance - investigating the negative consequences ofchatgpt-assisted problem solving in physics education}.
\newblock \emph{Preprint}, arXiv:2309.03087.

\bibitem[{Lee et~al.(2023)Lee, Phatale, Mansoor, Lu, Mesnard, Bishop, Carbune, and Rastogi}]{RLAIF}
Harrison Lee, Samrat Phatale, Hassan Mansoor, Kellie Lu, Thomas Mesnard, Colton Bishop, Victor Carbune, and Abhinav Rastogi. 2023.
\newblock \href {https://arxiv.org/abs/2307.04964} {Rlaif: Scaling reinforcement learning from human feedback with ai feedback}.
\newblock \emph{Preprint}, arXiv:2307.04964.

\bibitem[{Li et~al.(2023{\natexlab{a}})Li, Xu, Zhang, Yu, Sun, and Luo}]{RHFWithReMax}
Ziniu Li, Tian Xu, Yushun Zhang, Yang Yu, Ruoyu Sun, and Zhi-Quan Luo. 2023{\natexlab{a}}.
\newblock \href {https://arxiv.org/abs/2310.10505} {Remax: A simple, effective, and efficient reinforcement learning method for aligning large language models}.
\newblock \emph{Preprint}, arXiv:2310.10505.

\bibitem[{Li et~al.(2023{\natexlab{b}})Li, Xu, Zhang, Yu, Sun, and Luo}]{li2023remax}
Ziniu Li, Tian Xu, Yushun Zhang, Yang Yu, Ruoyu Sun, and Zhi-Quan Luo. 2023{\natexlab{b}}.
\newblock \href {https://arxiv.org/abs/2310.10505} {Remax: A simple, effective, and efficient reinforcement learning method for aligning large language models}.
\newblock \emph{Preprint}, arXiv:2310.10505.

\bibitem[{Lin(2004)}]{lin-2004-rouge}
Chin-Yew Lin. 2004.
\newblock \href {https://aclanthology.org/W04-1013} {{ROUGE}: A package for automatic evaluation of summaries}.
\newblock In \emph{Text Summarization Branches Out}, pages 74--81, Barcelona, Spain. Association for Computational Linguistics.

\bibitem[{Liu et~al.(2022)Liu, Tam, Muqeeth, Mohta, Huang, Bansal, and Raffel}]{liu2022fewshot}
Haokun Liu, Derek Tam, Mohammed Muqeeth, Jay Mohta, Tenghao Huang, Mohit Bansal, and Colin Raffel. 2022.
\newblock \href {https://arxiv.org/abs/2205.05638} {Few-shot parameter-efficient fine-tuning is better and cheaper than in-context learning}.
\newblock \emph{Preprint}, arXiv:2205.05638.

\bibitem[{Luo et~al.(2023)Luo, Sun, Xu, Zhao, Lou, Tao, Geng, Lin, Chen, and Zhang}]{luo2023wizardmath}
Haipeng Luo, Qingfeng Sun, Can Xu, Pu~Zhao, Jianguang Lou, Chongyang Tao, Xiubo Geng, Qingwei Lin, Shifeng Chen, and Dongmei Zhang. 2023.
\newblock \href {https://arxiv.org/abs/2308.09583} {Wizardmath: Empowering mathematical reasoning for large language models via reinforced evol-instruct}.
\newblock \emph{Preprint}, arXiv:2308.09583.

\bibitem[{Nakano et~al.(2023)Nakano, Hilton, Balaji, Wu, Ouyang, Kim, Hesse, Jain, Kosaraju, and Saunders}]{webgpt}
Reiichiro Nakano, Jacob Hilton, Suchir Balaji, Jeff Wu, Long Ouyang, Christina Kim, Christopher Hesse, Shantanu Jain, Vineet Kosaraju, and William Saunders. 2023.
\newblock \href {https://arxiv.org/abs/2112.09332} {Webgpt: Browser-assisted question-answering with human feedback}.
\newblock \emph{Preprint}, arXiv:2112.09332.

\bibitem[{Ouyang et~al.(2022{\natexlab{a}})Ouyang, Wu, Jiang, Almeida, Wainwright, Mishkin, Zhang, Agarwal, Slama, Ray, Schulman, Hilton, Kelton, Miller, Simens, Askell, Welinder, Christiano, Leike, and Lowe}]{ouyang2022training}
Long Ouyang, Jeff Wu, Xu~Jiang, Diogo Almeida, Carroll~L. Wainwright, Pamela Mishkin, Chong Zhang, Sandhini Agarwal, Katarina Slama, Alex Ray, John Schulman, Jacob Hilton, Fraser Kelton, Luke Miller, Maddie Simens, Amanda Askell, Peter Welinder, Paul Christiano, Jan Leike, and Ryan Lowe. 2022{\natexlab{a}}.
\newblock \href {https://arxiv.org/abs/2203.02155} {Training language models to follow instructions with human feedback}.
\newblock \emph{Preprint}, arXiv:2203.02155.

\bibitem[{Ouyang et~al.(2022{\natexlab{b}})Ouyang, Wu, Jiang, Almeida, Wainwright, Mishkin, Zhang, Agarwal, Slama, Ray, Schulman, Hilton, Kelton, Miller, Simens, Askell, Welinder, Christiano, Leike, and Lowe}]{NEURIPS2022_b1efde53}
Long Ouyang, Jeffrey Wu, Xu~Jiang, Diogo Almeida, Carroll Wainwright, Pamela Mishkin, Chong Zhang, Sandhini Agarwal, Katarina Slama, Alex Ray, John Schulman, Jacob Hilton, Fraser Kelton, Luke Miller, Maddie Simens, Amanda Askell, Peter Welinder, Paul~F Christiano, Jan Leike, and Ryan Lowe. 2022{\natexlab{b}}.
\newblock Training language models to follow instructions with human feedback.

\bibitem[{Papineni et~al.(2002)Papineni, Roukos, Ward, and Zhu}]{papineni-etal-2002-bleu}
Kishore Papineni, Salim Roukos, Todd Ward, and Wei-Jing Zhu. 2002.
\newblock \href {https://doi.org/10.3115/1073083.1073135} {{B}leu: a method for automatic evaluation of machine translation}.
\newblock In \emph{Proceedings of the 40th Annual Meeting of the Association for Computational Linguistics}, pages 311--318, Philadelphia, Pennsylvania, USA. Association for Computational Linguistics.

\bibitem[{Rafailov et~al.(2023)Rafailov, Sharma, Mitchell, Ermon, Manning, and Finn†}]{RHFWithDPO}
Rafael Rafailov, Archit Sharma, Eric Mitchell, Stefano Ermon, Christopher~D. Manning, and Chelsea Finn†. 2023.
\newblock \href {https://arxiv.org/abs/2305.18290} {Direct preference optimization: Your language model is secretly a reward model}.
\newblock \emph{Preprint}, arXiv:2305.18290.

\bibitem[{Schulman et~al.(2017)Schulman, Wolski, Dhariwal, Radford, and Klimov}]{schulman2017proximal}
John Schulman, Filip Wolski, Prafulla Dhariwal, Alec Radford, and Oleg Klimov. 2017.
\newblock \href {https://arxiv.org/abs/1707.06347} {Proximal policy optimization algorithms}.
\newblock \emph{Preprint}, arXiv:1707.06347.

\bibitem[{Stiennon et~al.(2023)Stiennon, Ouyang, Wu, Ziegler, Lowe, Voss, and Christiano}]{LearningSHF}
Nisan Stiennon, Long Ouyang, Jeff Wu, Daniel~M. Ziegler, Ryan Lowe, Chelsea Voss, and Alec Radford Dario Amodei~Paul Christiano. 2023.
\newblock \href {https://arxiv.org/abs/2009.01325} {Learning to summarize from human feedback}.
\newblock \emph{Preprint}, arXiv:2009.01325.

\bibitem[{Team et~al.(2024)Team, Anil, Borgeaud, Alayrac, and et. al}]{geminiteam2024gemini}
Gemini Team, Rohan Anil, Sebastian Borgeaud, Jean-Baptiste Alayrac, and Jiahui~Yu et. al. 2024.
\newblock \href {https://arxiv.org/abs/2312.11805} {Gemini: A family of highly capable multimodal models}.
\newblock \emph{Preprint}, arXiv:2312.11805.

\bibitem[{Touvron et~al.(2023)Touvron, Martin, Stone, Albert, Almahairi, Babaei, Bashlykov, Batra, Bhargava, Bhosale, Bikel, Blecher, Ferrer, Chen, Cucurull, Esiobu, Fernandes, Fu, Fu, Fuller, Gao, Goswami, Goyal, Hartshorn, Hosseini, Hou, Inan, Kardas, Kerkez, Khabsa, Kloumann, Korenev, Koura, Lachaux, Lavril, Lee, Liskovich, Lu, Mao, Martinet, Mihaylov, Mishra, Molybog, Nie, Poulton, Reizenstein, Rungta, Saladi, Schelten, Silva, Smith, Subramanian, Tan, Tang, Taylor, Williams, Kuan, Xu, Yan, Zarov, Zhang, Fan, Kambadur, Narang, Rodriguez, Stojnic, Edunov, and Scialom}]{touvron2023llama}
Hugo Touvron, Louis Martin, Kevin Stone, Peter Albert, Amjad Almahairi, Yasmine Babaei, Nikolay Bashlykov, Soumya Batra, Prajjwal Bhargava, Shruti Bhosale, Dan Bikel, Lukas Blecher, Cristian~Canton Ferrer, Moya Chen, Guillem Cucurull, David Esiobu, Jude Fernandes, Jeremy Fu, Wenyin Fu, Brian Fuller, Cynthia Gao, Vedanuj Goswami, Naman Goyal, Anthony Hartshorn, Saghar Hosseini, Rui Hou, Hakan Inan, Marcin Kardas, Viktor Kerkez, Madian Khabsa, Isabel Kloumann, Artem Korenev, Punit~Singh Koura, Marie-Anne Lachaux, Thibaut Lavril, Jenya Lee, Diana Liskovich, Yinghai Lu, Yuning Mao, Xavier Martinet, Todor Mihaylov, Pushkar Mishra, Igor Molybog, Yixin Nie, Andrew Poulton, Jeremy Reizenstein, Rashi Rungta, Kalyan Saladi, Alan Schelten, Ruan Silva, Eric~Michael Smith, Ranjan Subramanian, Xiaoqing~Ellen Tan, Binh Tang, Ross Taylor, Adina Williams, Jian~Xiang Kuan, Puxin Xu, Zheng Yan, Iliyan Zarov, Yuchen Zhang, Angela Fan, Melanie Kambadur, Sharan Narang, Aurelien Rodriguez, Robert Stojnic, Sergey Edunov, and Thomas
  Scialom. 2023.
\newblock \href {https://arxiv.org/abs/2307.09288} {Llama 2: Open foundation and fine-tuned chat models}.
\newblock \emph{Preprint}, arXiv:2307.09288.

\bibitem[{Vaswani et~al.(2023)Vaswani, Shazeer, Parmar, Uszkoreit, Jones, Gomez, Kaiser, and Polosukhin}]{vaswani2023attention}
Ashish Vaswani, Noam Shazeer, Niki Parmar, Jakob Uszkoreit, Llion Jones, Aidan~N. Gomez, Lukasz Kaiser, and Illia Polosukhin. 2023.
\newblock \href {https://arxiv.org/abs/1706.03762} {Attention is all you need}.
\newblock \emph{Preprint}, arXiv:1706.03762.

\bibitem[{Xie et~al.(2023)Xie, Xie, Lin, Wei, Li, Kong, Chen, Zhuo, Hu, and Li}]{OlaGPT}
Yuanzhen Xie, Tao Xie, Mingxiong Lin, WenTao Wei, Chenglin Li, Beibei Kong, Lei Chen, Chengxiang Zhuo, Bo~Hu, and Zang Li. 2023.
\newblock \href {https://arxiv.org/abs/2305.16334} {Olagpt: Empowering llms with human-like problem-solving abilities}.
\newblock \emph{Preprint}, arXiv:2305.16334.

\bibitem[{Xu et~al.(2024)Xu, Fu, Gao, Ye, Liu, Mei, Wang, Yu, and Wu}]{xu2024dpo}
Shusheng Xu, Wei Fu, Jiaxuan Gao, Wenjie Ye, Weilin Liu, Zhiyu Mei, Guangju Wang, Chao Yu, and Yi~Wu. 2024.
\newblock \href {https://arxiv.org/abs/2404.10719} {Is dpo superior to ppo for llm alignment? a comprehensive study}.
\newblock \emph{Preprint}, arXiv:2404.10719.

\bibitem[{Yu et~al.(2023)Yu, Jiang, Shi4, Yu, Liu, Zhang, Kwok, Li, Weller, and Liu1}]{metamath}
Longhui Yu, Weisen Jiang, Han Shi4, Jincheng Yu, Zhengying Liu, Yu~Zhang, James~T. Kwok, Zhenguo Li, Adrian Weller, and Weiyang Liu1. 2023.
\newblock \href {https://arxiv.org/abs/2309.12284} {Metamath: Bootstrap your own mathematical questions for large language models}.
\newblock \emph{Preprint}, arXiv:2309.12284.

\bibitem[{Yue et~al.(2023)Yue, Qu, Zhang, Fu, Huang, Sun, Su, and Chen}]{yue2023mammoth}
Xiang Yue, Xingwei Qu, Ge~Zhang, Yao Fu, Wenhao Huang, Huan Sun, Yu~Su, and Wenhu Chen. 2023.
\newblock \href {https://arxiv.org/abs/2309.05653} {Mammoth: Building math generalist models through hybrid instruction tuning}.
\newblock \emph{Preprint}, arXiv:2309.05653.

\bibitem[{Zhang et~al.(2020)Zhang, Kishore, Wu, Weinberger, and Artzi}]{zhang2020bertscore}
Tianyi Zhang, Varsha Kishore, Felix Wu, Kilian~Q. Weinberger, and Yoav Artzi. 2020.
\newblock \href {https://arxiv.org/abs/1904.09675} {Bertscore: Evaluating text generation with bert}.
\newblock \emph{Preprint}, arXiv:1904.09675.

\bibitem[{Zheng et~al.(2023)Zheng, Dou, Gao, Hua, Shen, Wang, Liu, Jin, Liu, Zhou, Xiong, Chen, Xi, Xu, Lai, Zhu, Chang, Yin, Weng, Cheng, Huang, Sun, Yan, Gui, Zhang, Qiu, and Huang}]{RHFWithPPO}
Rui Zheng, Shihan Dou, Songyang Gao, Yuan Hua, Wei Shen, Binghai Wang, Yan Liu, Senjie Jin, Qin Liu, Yuhao Zhou, Limao Xiong, Lu~Chen, Zhiheng Xi, Nuo Xu, Wenbin Lai, Minghao Zhu, Cheng Chang, Zhangyue Yin, Rongxiang Weng, Wensen Cheng, Haoran Huang, Tianxiang Sun, Hang Yan, Tao Gui, Qi~Zhang, Xipeng Qiu, and Xuanjing Huang. 2023.
\newblock \href {https://arxiv.org/abs/2307.04964} {Secrets of rlhf in large language models part i: Ppo}.
\newblock \emph{Preprint}, arXiv:2307.04964.

\bibitem[{Ziegler et~al.(2020)Ziegler, Stiennon, Wu, Brown, Radford, Amodei, Christiano, and Irving}]{ziegler2020finetuning}
Daniel~M. Ziegler, Nisan Stiennon, Jeffrey Wu, Tom~B. Brown, Alec Radford, Dario Amodei, Paul Christiano, and Geoffrey Irving. 2020.
\newblock \href {https://arxiv.org/abs/1909.08593} {Fine-tuning language models from human preferences}.
\newblock \emph{Preprint}, arXiv:1909.08593.

\end{thebibliography}




\end{document}